\documentclass[11pt,a4paper]{article}
\usepackage{float}
\usepackage[hyperref]{acl2019}
\usepackage{times}
\usepackage{latexsym}
\usepackage{multirow}
\usepackage{graphicx}
\usepackage{amsmath}
\usepackage{comment}
\graphicspath{ {images/} }

\usepackage{url}

\aclfinalcopy 

\usepackage{flushend}

\title{Multilingual, Multi-scale and Multi-layer Visualization of \\ Intermediate Representations} 

\author{Carlos Escolano$^{*\diamond}$, Marta R. Costa-juss\`a${^{*\diamond}}$, Elora Lacroux$^\diamond$ and Pere-Pau V\'azquez$^{\star\diamond}$ \\
$^*$ TALP Research Center, $^\diamond$Universitat Polit\`ecnica de Catalunya, Barcelona \\
$^\star$ ViRVIG Group \\
  \texttt{\{carlos.escolano,marta.ruiz\}@upc.edu}\\\texttt{lacrouxelora@gmail.com,pere.pau@cs.upc.edu }}

\date{}

\begin{document}
\maketitle

\begin{abstract}
The main alternatives nowadays to deal with sequences are Recurrent Neural Networks (RNN), Convolutional Neural Networks (CNN) architectures and the Transformer. 
In this context, RNN's, CNN's and Transformer have most commonly been used as an encoder-decoder architecture with multiple layers in each module. Far beyond this, these architectures are the basis for the contextual word embeddings which are revolutionizing most natural language downstream applications. 

However, intermediate layer representations in sequence-based architectures can be difficult to interpret. To make each layer representation within these architectures more accessible and meaningful, we introduce a web-based tool that visualizes them both at the sentence and token level. We present three use cases. The first analyses gender issues in contextual word embeddings. The second and third are showing multilingual intermediate representations for sentences and tokens and the evolution of these intermediate representations along the multiple layers of the decoder and in the context of multilingual machine translation.
\end{abstract}

\section{Introduction}

The Transformer \cite{vaswani2017attention} is a powerful architecture that was initially proposed to train neural machine translation. 
This architecture 
deals with variable sequences by concatenating feed-forward networks and attention-based mechanisms. While the composed modules of the Transformer may not be complex by themselves, it is the composition of several layers of these modules that make the entire architecture less interpretable.



We are aiming at providing a tool to give insights to the sentences and token representation from each layer in the Transformer. Far beyond the Transfomer interpretation which has become by de-facto the state-of-the-art in machine translation, our tool is able to represent intermediate representations of other sequence-based architectures such as RNNs \cite{bahdanau2014neural} or ConvS2S \cite{Gehring:2017:CSS:3305381.3305510} as well. Note that sequence-based architectures are having impact in many multimodal applications such as image captioning and speech recognition \cite{kaiser2017one,44926}.

The uses of our visualization tool are quite a few varying from social bias, multilingual or linguistic analysis. In particular, we focus in analysing the gender inequalities in contextual word embeddings and the common language representation in a multilingual machine translation system. 







\section{Visualization tool}

In this section we present a multi-scale and multi-layer visualization tool for the sequence-based architectures, available as tool\footnote{https://github.com/elorala/interlingua-visualization} and as a demo\footnote{https://upc-nmt-vis.herokuapp.com/}. 
The tool is implemented in Python using the Bokeh library for data visualization and the Flask library as web microfamework to embed the Bokeh dashboards on the webpage.

The tool consists in using as input fixed-representations, being a matrix of dimensions the embedding size per sentence length (in tokens). Therefore, the input data required are the sentences to be represented (txt), the sentence representations (json) and optionally the tokens embeddings (json). Then, a UMAP \cite{mcinnes2018umap-software} dimensionality reduction is performed to plot the representation of this multidimensional data in two dimensions. This dimensionality reduction is performed for the fixed-representations at the sentence and token level. The tool comprises two views: multi-scale intermediate representation for one layer and multi-layer sentence representation. These two views can be either monolingual or multilingual. The main page of the tool comprises these two views for the user to choose.

We describe these two views on different use cases. For the first view, we show the use cases of detection of gender bias in contextual word embeddings and common representation in multilingual machine translation. For the second view, the use case builds on layer interpretation of multi-way parallel sentences in a translation decoder and showing which layer carries out higher semantic meaning. 


\subsection{Multi-scale Intermediate representation}

This visualization consists on two coordinated views, that encode different information through scatterplots.  
The one on the left shows the M sentence intermediate representations. 
Each dot in the sentence graph corresponds to one sentence, 
by hovering on a point we visualize the sentence as well as the arrows to the corresponding translation sentences, in case we are working with multilingual data. 
There is an option to visualize a particular sentence by writing it in the search bar. The search bar has an autocomplete feature (activated when typing two characters) and then, the user can click on the right suggestion. 

The right view shows the tokens. Initially, when no sentence from the previous view is selected, this plot shows all vocabulary tokens. By brushing over one or more sentences (in left view), the right view filters out the tokens \emph{not belonging} to the selected sentence (and the tokens that compose the parallel sentences in the other languages). Once the user selects a sentence by clicking or searching, only the words from this sentence (and its translations) remain on the chart. By hovering on a point, the user can see the text of the word, analogously to the sentences view. 

\begin{figure*}[h]
\begin{center}
  \includegraphics[width=0.85\textwidth]{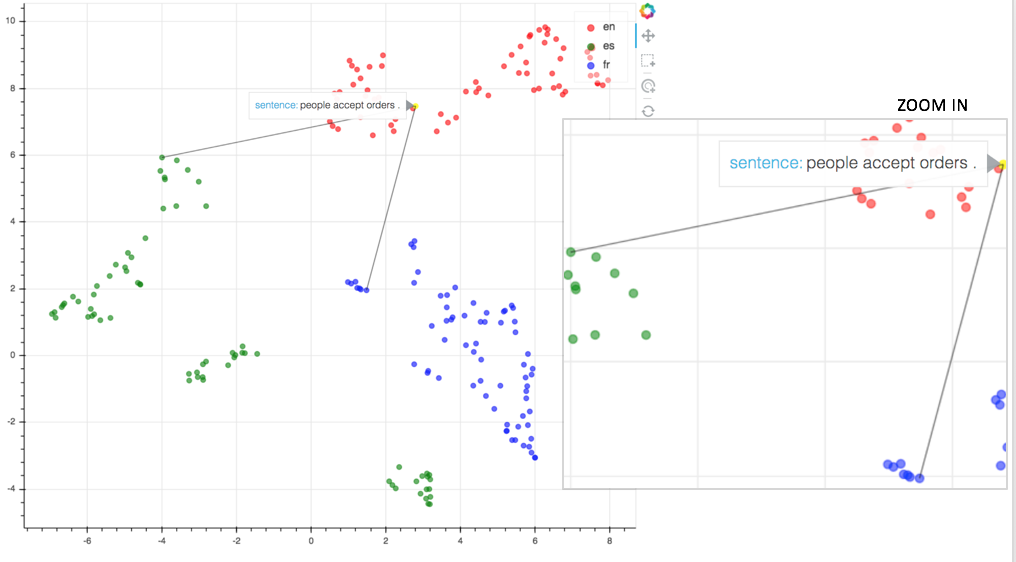}
  \caption{Multilingual common representation at the sentence level (sentence \textit{people accept orders .})}\label{fig:multilingualsentences}
 \end{center}
\end{figure*}

\begin{figure}[h]
\begin{center}
  \includegraphics[width=0.49\textwidth]{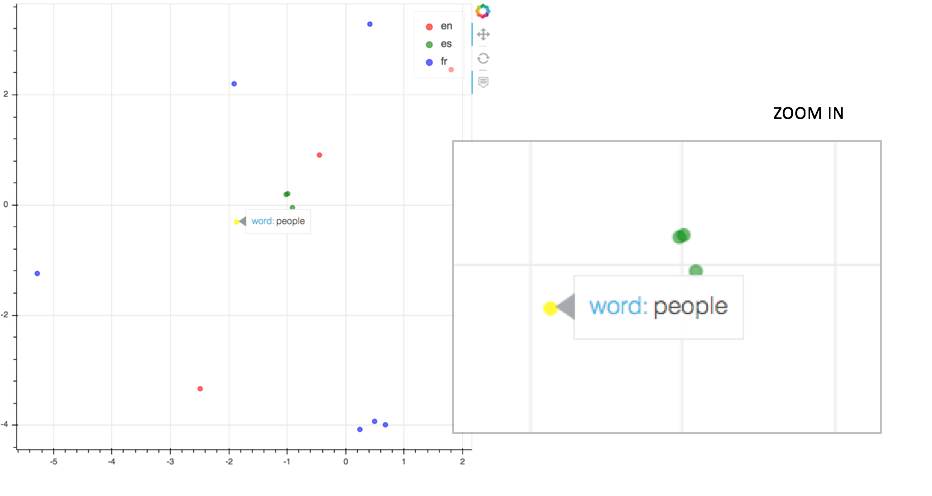}
  \caption{Multilingual common representation at the token level (token \textit{people})}\label{fig:multilingualwords}
 \end{center}
\end{figure}

Sentences and tokens can be simultaneously visualised for all languages that we are studying and we can interpret the intermediate representation in terms of both granularity levels.
See Figures \ref{fig:multilingualsentences} and \ref{fig:multilingualwords} for illustration, which are as well examples of the second use case (explained as follows).



\paragraph{Use case 1: Gender bias in Contextual Word Embeddings.}

\begin{figure}[h]
\begin{center}
  \includegraphics[width=0.45\textwidth]{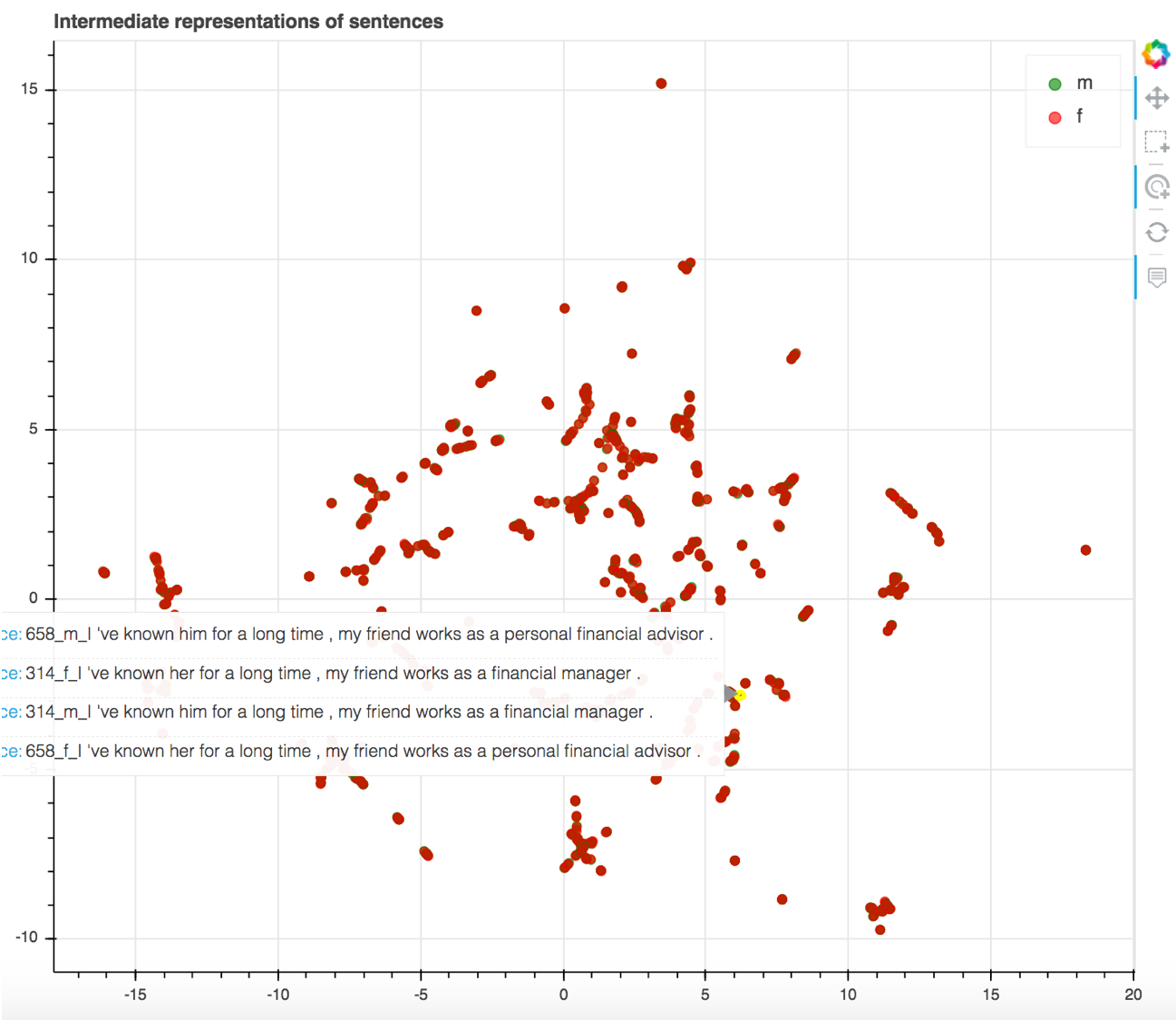}
  \caption{Contextual word embedding representation at the sentence level (sentences \textit{I've known him/her for a long time, my friend is a financial manager/personal financial advisor})\label{fig:gb3}}
 \end{center}
\end{figure}

\begin{figure}[h]
\begin{center}
  \includegraphics[width=0.43\textwidth]{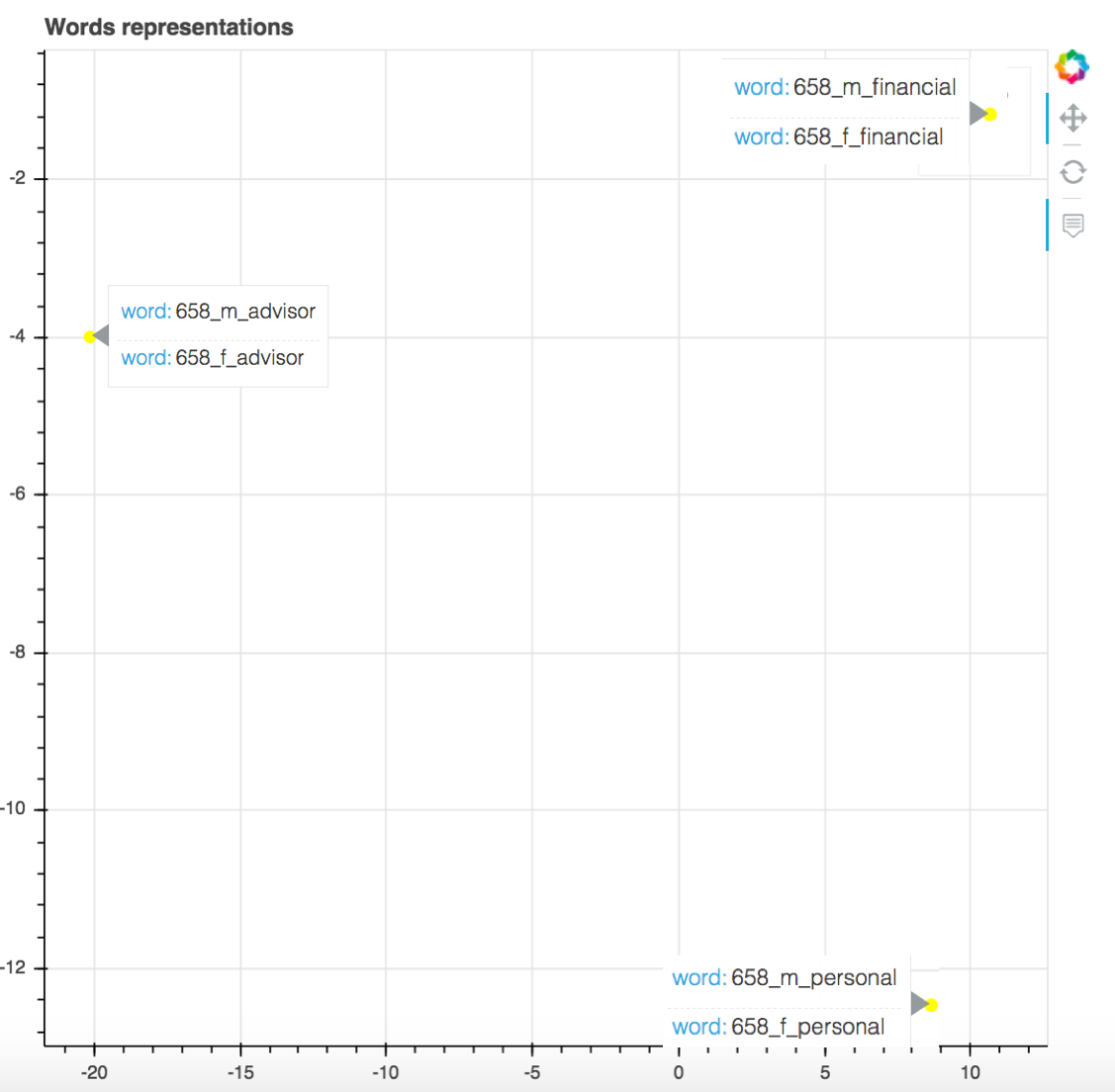}
  \caption{Contextual word embedding representation at the token level \textit{financial manager}}\label{fig:gb1}
 \end{center}
\end{figure}

\begin{figure}[h]
\begin{center}
  \includegraphics[width=0.43\textwidth]{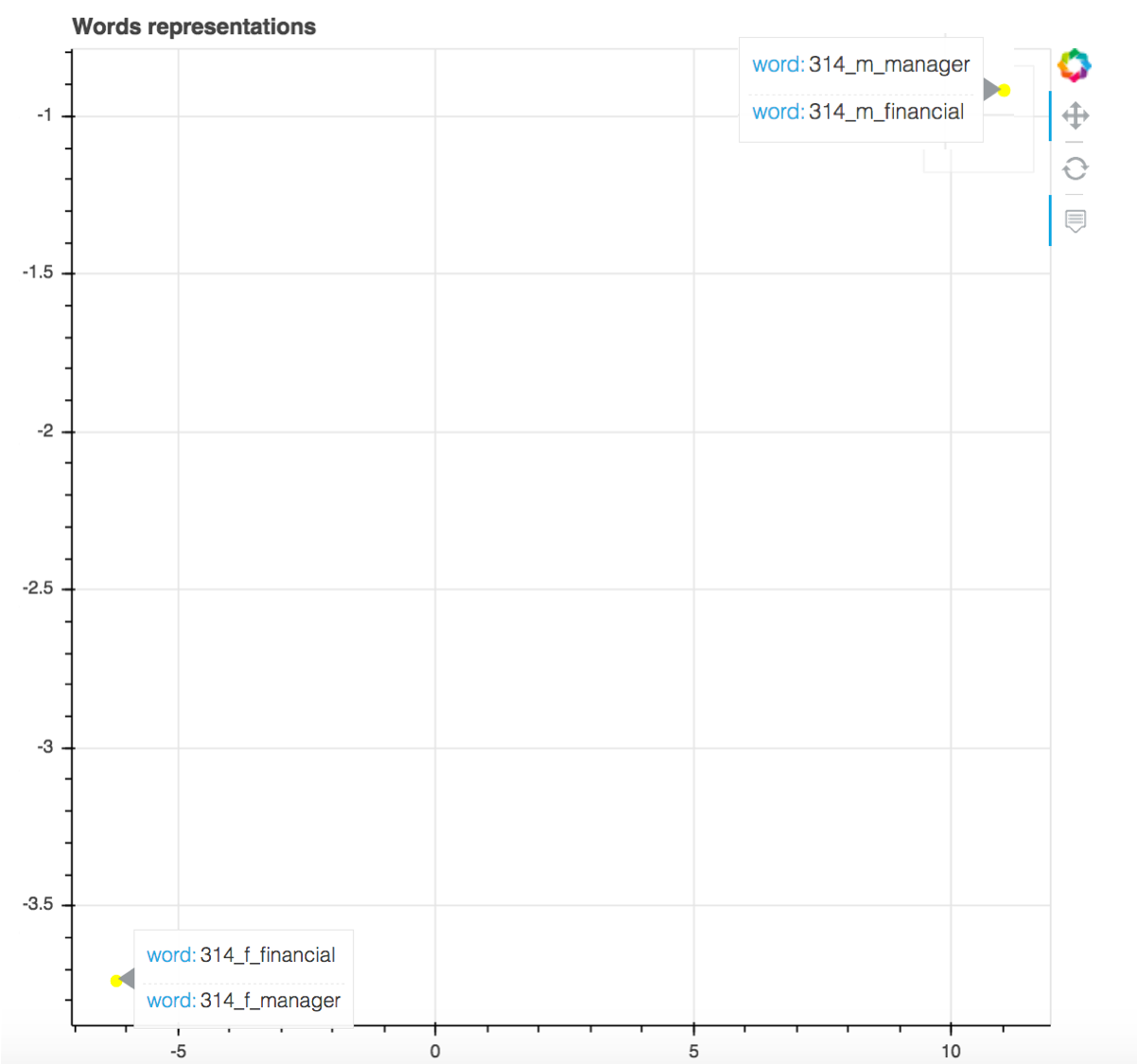}
  \caption{Contextual word embedding representation at the token level \textit{personal financial advisor}}\label{fig:gb2}
 \end{center}
\end{figure}

The objective of this use case is to visualize the contextual word representations on a set of occupational vocabulary. We use the ELMO implementation \cite{elmo}, based on RNNs and as data, we use 1019 sentences from previous work \cite{font:2019} that follow the next template \textit{I've known \textit{him/her} for a long time, my friend works as a \textit{occupation}}. Examples of occupations include: \textit{accounting clerk, nurse midwife or biological scientist}. Since we have two sets: one for female templates and another for male templates, we use the two sets as if they were different languages. We visualize sentences and words. For sentences, we see that sentences with similar professions (i.e. \textit{financial manager}, \textit{personal financial advisor}) tend to be close in the space for both female and male versions. See Figure \ref{fig:gb3}. However, when visualizing the word representations, in the case of \textit{financial manager}, words for female and male representation are placed in very distant points in the space as seen in Figure \ref{fig:gb1}. On the contrary, words for female and male representation in the case of \textit{personal financial advisor} are represented together as seen in Figure \ref{fig:gb2}. So, we conclude that \textit{financial} in a male/female context is differently represented if attached to \textit{manager} but the same \textit{financial} is similarly represented in male/female context if attached to \textit{personal} and \textit{advisor}. Our tool allows to visualize that contextual word embeddings encode gender biases and this conclusion is coherent with previous literature experiments \cite{basta:2019}.

\paragraph{Use case 2: Multilingual common representation in translation.}

Nowadays, there are two main architectures for multilingual neural machine translation which are a universal shared encoder and decoder and independent multiple encoders and decoders. In both cases, there is an intermediate representation where sentences that have similar meanings should be represented close in the space. For our second and third use case, we use the intermediate representations of the multilingual Transformer-based architecture presented in \cite{escolano:2019}. Basically, the architecture consists in independent encoders and decoders with a forced-interlingua space. This system is trained on data extracted from the UN \cite{ziemski2016united} and EPPS datasets \cite{koehn2005europarl} that provide 15 million parallel sentences between English and Spanish and French. \textit{newstest2012} and \textit{newstest2013} were used as validation and test sets, respectively. These sets provide parallel data between the three languages. 



Figure \ref{fig:multilingualsentences} shows 130 sentences extracted from the test set, in the \textit{3} languages at hand  and in the common space (at the output of the encoder). When we select a particular sentence (e.g. \textit{people accept orders .}), for each token in the sentence selected, the user can select to visualize the token representations (e.g. \textit{people}) as shown in Figure \ref{fig:multilingualwords}. 

\subsection{Multi-layer sentence representation}

\begin{figure*}[h!]
\begin{center}
  \includegraphics[width=0.81\textwidth]{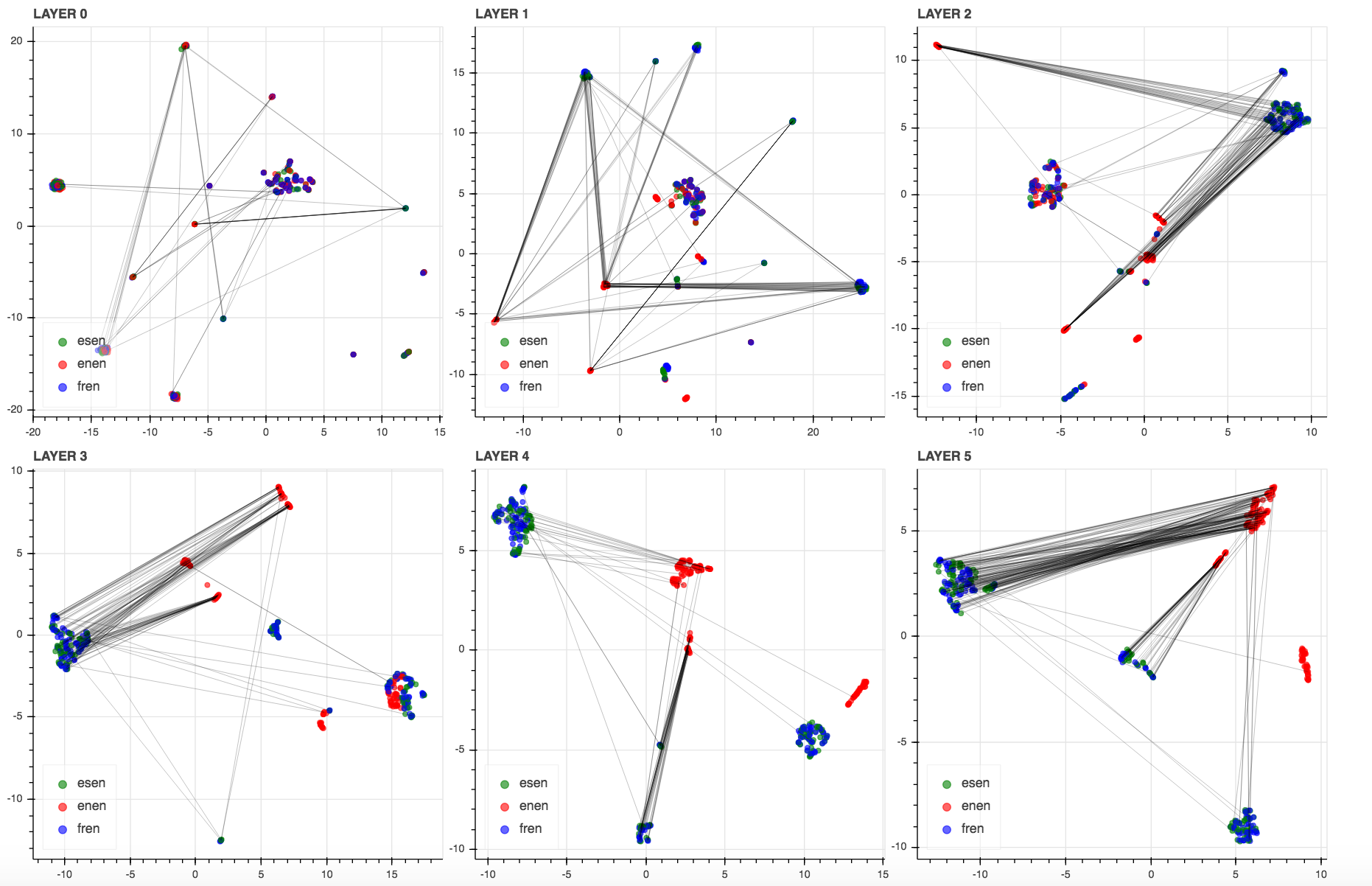}
  \caption{Decoder layer multilingual sentence representation}\label{fig:multilinguallayers}
 \end{center}
\end{figure*}

This visualization shows T layers simultaneously for single or multiple languages in a small multiples design. This facilitates the analysis of sentence representation evolution across all the layers of the Transformer at once. See Figure \ref{fig:multilinguallayers}.

On each view, we can display the sentence by hovering. In order to emphasize the distances between the translations and to have a better insight of the evolution, the link between the most dissimilar are displayed on the plots. By hovering on the lines, the user can obtain the cosine distance value computed on SciPy.
On the views, only the distances superior to 1 are displayed. Even if the dimensionality reduction of UMAP does show interpretable distances \cite{mcinnes2018umap-software}, showing consecutive layers of the Transformer, and seeing the evolution of the representations allows us to draw hints about the layer roles as we will see in the third use case.

Finally, the tool allows for analysis in multiple layers and languages. This means that initially, the multiple layers represented on the dashboard are in one particular language. However, the user can switch to the multiple layers from another language by using the selection tool at the top of the page. Since all views are synchronized, upon changing the language set, all of them change accordingly.

\paragraph{Use case 3: Multilingual Layer Interpretation in Translation Decoding} Encoders and decoders in a neural machine translation system are usually composed of different layers. The role of each layer is difficult to interpret. Visualizing sentences at each of these layers can help us on identifying the sentence distance evolution giving us hints of different linguistic roles for the layers when compared between them. 

\begin{figure*}[h!]
\begin{center}
  \includegraphics[width=0.8\textwidth]{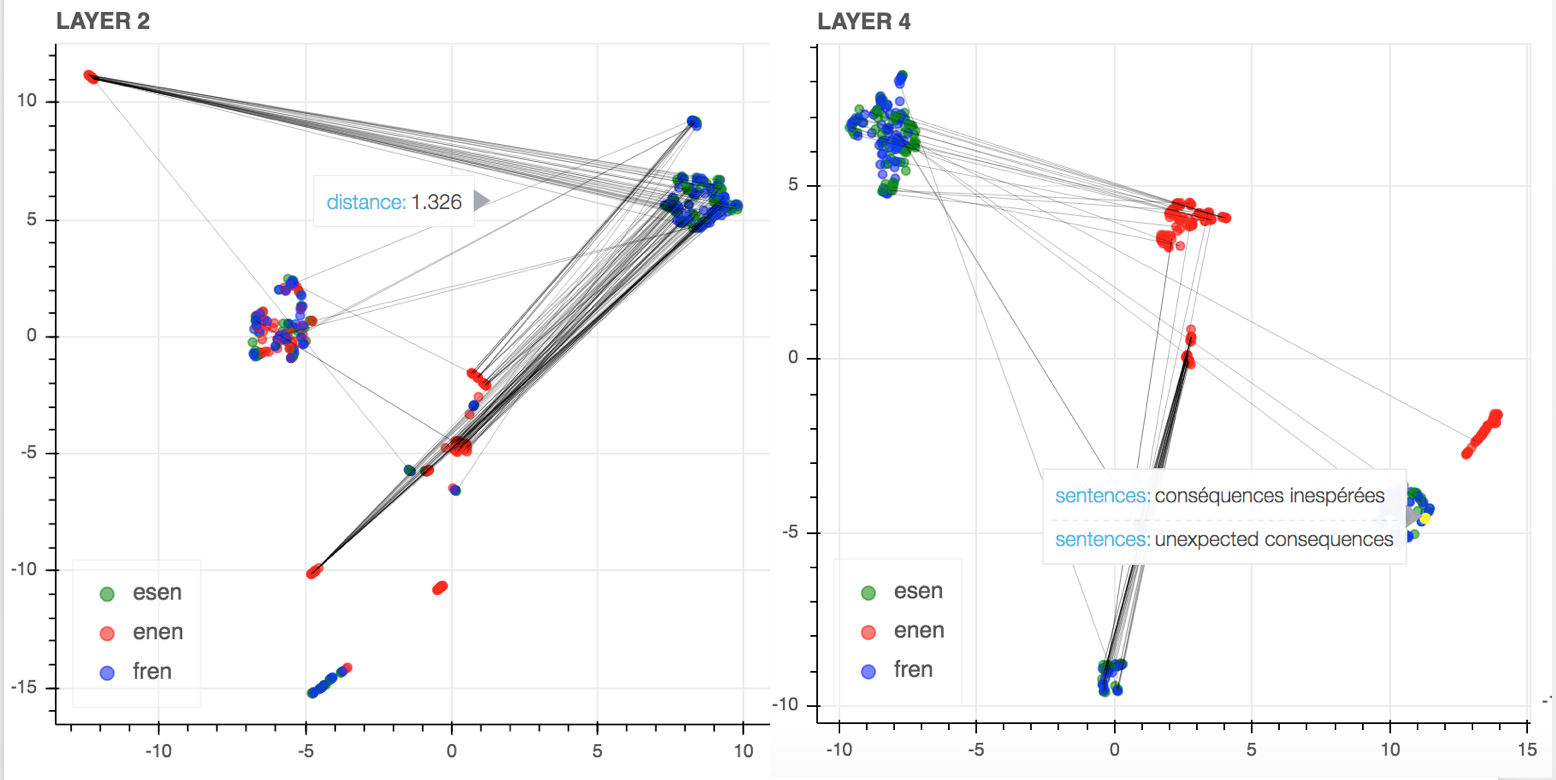}
  \caption{Decoder layer multilingual sentence representation: distance and sentences}\label{fig:multilinguallayers24}
 \end{center}
\end{figure*}

In the current example, we are representing the same set and architecture as in user case 2 but for the 6 decoder layers. Figure \ref{fig:multilinguallayers} shows the plot for the six layers and Figure \ref{fig:multilinguallayers24} shows how it performs hovering on a point (e.g. showing sentences, right) and hovering on a line (e.g. showing distance measure, left), respectively. 

Since we show sentences with the same meaning and in different languages, we can interpret that the layer that tends to better cluster the parallel sentence than in previous or subsequent layers can be interpreted as the layer with higher semantic implications. From Figure \ref{fig:multilinguallayers} note that higher layers in the decoder (specially 4 and 5) group sentences together more than previous layers (see reference axes in Figure \ref{fig:multilinguallayers}).

\section{Adaptability}

\begin{figure}[h]
\begin{minipage}[b]{0.45\linewidth}
\centering
\includegraphics[scale=0.42]{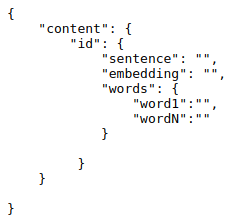}

\end{minipage}
\hspace{0.5cm}
\begin{minipage}[b]{0.45\linewidth}
\centering
\includegraphics[scale=0.42]{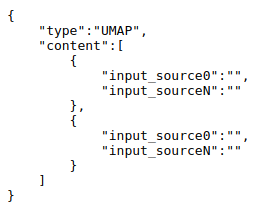}
\end{minipage}

\caption{Required JSON structures: (Left) use cases 1 and 2 and (Right) use case 3}
\label{fig:json}
\end{figure}

In this paper, we have discussed three use cases. However, our tool is highly flexible and adaptable, and and it allows for a large variety of tasks. 
The system only requires data to be formatted as a \textit{JSON} file following the structures defined in Figure \ref{fig:json}.  

The structure from use cases 1 and 2 defines the relation between sentence and token representations. For each token and embedding  a 2-dimensional is defined, showing its coordinates in the final plots. 

On the other side, the structure  from use case 3 contains the representations of the layers to be plotted and it is described as an array containing the coordinates for each sentence.

This implementation allows our tool to be agnostic to factors such as vocabulary sizes and dimensionality reductions techniques, as they they are applied before \textit{JSON} creation.

\section{Related Work}

Given the versatility of the sequence architectures, current tool feeds and relates to vast areas of research including contextual word embeddings, multilingual models, visualization and interpretability of sequence models, zero-shot learning. However, we just refer here to closest and recent works. Regarding related demonstrations, authors in \cite{vig:2019} analyse the attention in the Transformer at multiple-scales and show different use cases on contextual word embeddings. Closest related work to our use cases is mentioned as follows.

\paragraph{Gender bias.} Gender bias has recently been analysed in contextual word embeddings \cite{zhao-etal-2019-gender,basta:2019}. Our tool aims at following-up this kind of research to work towards techniques that are able to neutralize these and other social biases.

\paragraph{Multilinguality analysis.} It is quite a common practice to visualize intermediate representations of sequence-to-sequence models \cite{johnson2017google,escolano:2019}. Our tool is not limited to this sentence representation of the intermediate representation, but it also includes the token-level representation. By simultaneously providing these two-granularity level representation we are aiming at a deeper analysis for both monolingual, cross-lingual and multilingual natural language processing downstream applications in general.

\paragraph{Linguistic insights.} \cite{raganato-tiedemann-2018-analysis} show interesting findings about
dependency relations and syntactic and semantic behavior across Transformer layers. Following this research line, our tool can further analyse how similar sentences in multiple languages evolve in their intermediate layer representations as well as monolingual sentences with same syntactic or morphological patterns.

\section{Conclusions}

We have presented an extremely flexible and adaptable visualization tool for multilingual intermediate representations of text both at the sentence and token's level. Together with our tool we have presented three use cases in the context of gender bias analysis in contextual word embeddings and for multilingual intermediate representations of machine translation.

\section*{Acknowledgements}

Authors want to thank Christine Raouf Basta for sharing her expertise in contextual word embeddings.
This work is supported in part by a Google Faculty Research Award.
This work is also supported in part by the
Spanish Ministerio de Econom\'{i}a y Competitividad,
the European Regional Development Fund
and the Agencia Estatal de Investigaci\'{o}n,
through the postdoctoral senior grant Ram\'{o}n y Cajal, contracts TEC2015-69266-P and TIN2017-88515-C2-1-R(GEN3DLIVE)
(MINECO/FEDER,EU), and contract PCIN-2017-079 (AEI/MINECO).

\bibliographystyle{acl_natbib}
\bibliography{acl2019}

\end{document}